\documentclass{article}
\usepackage[preprint]{log_2025}			

\usepackage{booktabs}						
\usepackage{multirow}						
\usepackage{amsfonts}						
\usepackage{graphicx}						
\usepackage{duckuments}						

\usepackage[numbers,compress,sort]{natbib}	

\usepackage{amsmath}
\usepackage{amssymb}
\usepackage{amsthm}
\usepackage{amsfonts}
\usepackage{mathtools}

\theoremstyle{plain}

\theoremstyle{definition}

\theoremstyle{remark}

\usepackage{algorithm}
\usepackage{algorithmic}
\usepackage[algo2e]{algorithm2e}
\usepackage[utf8]{inputenc}
\usepackage[cjk]{kotex}
\usepackage{wrapfig}
\usepackage{subfigure}
\usepackage{multirow}
\usepackage{pbox}
\usepackage{graphicx}
\usepackage{bm}
\usepackage{siunitx}
\usepackage{booktabs}
\usepackage{lipsum}
\usepackage{caption}
\usepackage{fixltx2e}
\usepackage{cancel}
\usepackage{colortbl}
\usepackage{enumitem}
\usepackage{nicefrac}
\usepackage{microtype}
\usepackage[toc,page,header]{appendix}
\usepackage{minitoc}

\usepackage[textsize=tiny]{todonotes}
\usepackage{marginnote}
 
\definecolor{bluegray}{rgb}{0.4, 0.6, 0.8}
\definecolor{electriclime}{rgb}{0.8, 1.0, 0.0}
\definecolor{malachite}{rgb}{0.04, 0.85, 0.32}
\definecolor{darkred}{rgb}{0.55, 0.0, 0.0}
\definecolor{darkblue}{rgb}{0.0, 0.0, 0.55}
\definecolor{darkgreen}{rgb}{0.0, 0.2, 0.13}
\definecolor{darkorchid}{rgb}{0.6, 0.2, 0.8}
\newcommand{\std}[1]{\tiny{\ensuremath{\pm}}\text{\tiny #1}}

\definecolor{lightRed}{HTML}{F8CECC}
\definecolor{redBorder}{HTML}{CC0000}
\definecolor{greenBorder}{HTML}{82B366}
\definecolor{lightGreen}{HTML}{D5E8D4}
\definecolor{blueBorder}{HTML}{6C8EBF}
\definecolor{lightBlue}{HTML}{DAE8FC}
\definecolor{blackBorder}{HTML}{000000}

\title{Can TabPFN Compete with GNNs for Node Classification via Graph Tabularization?}

\author[Choi et al.]{%
Jeongwhan Choi\\
KAIST \\
\email{jeongwhan.choi@kaist.ac.kr}\And
Woosung Kang\\
KAIST\\
\email{wskang@kaist.ac.kr}\And
Minseo Kim\\
KAIST\\
\email{evlingbling@kaist.ac.kr}\And
Jongwoo Kim\\
KAIST\\
\email{gsds4885@kaist.ac.kr}\And
Noseong Park\\
KAIST\\
\email{noseong@kaist.ac.kr}
}

\begin{document}

\maketitle

\begin{abstract}
Foundation models pretrained on large data have demonstrated remarkable zero-shot generalization capabilities across domains. Building on the success of TabPFN for tabular data and its recent extension to time series, we investigate whether graph node classification can be effectively reformulated as a tabular learning problem. We introduce TabPFN-GN, which transforms graph data into tabular features by extracting node attributes, structural properties, positional encodings, and optionally smoothed neighborhood features. This enables TabPFN to perform direct node classification without any graph-specific training or language model dependencies. Our experiments on 12 benchmark datasets reveal that TabPFN-GN achieves competitive performance with GNNs on homophilous graphs and consistently outperforms them on heterophilous graphs. These results demonstrate that principled feature engineering can bridge the gap between tabular and graph domains, providing a practical alternative to task-specific GNN training and LLM-dependent graph foundation models.
\end{abstract}

\section{Introduction}
Large-scale pretrained models trained on large datasets, such as foundation and large language models (LLMs)~\citep{brown2020language}, have gained popularity across diverse domains, including text~\citep{radford2019gpt2,lan2019albert,lewis2020bart}, images~\citep{oquab2023dinov2,bai2024sequential}, and time series~\citep{das2024timefm,liu2024moirai,auer2025tirex}, due to their ability to make accurate predictions with minimal fine-tuning on specific datasets. 
There have been recent efforts to build graph foundation models that take a seemingly natural approach to leveraging LLMs~\citep{chen2024text,chen2024llaga,li2024glbench,tang2024graphgpt}.
Meanwhile, a similar paradigm but a different approach has been proposed in the tabular domain. TabPFNs~\citep{hollmann2023tabpfn,hollmann2025tabpfnv2}, trained only on synthetic data generating numerical and categorical features, achieve remarkable performance on tabular tasks without fine-tuning or retraining on the target dataset. 
This success, particularly its recent extension to time series~\citep{hoo2024tabular,hoo2025tables}, suggests unexplored potential for graph learning. Graph neural networks (GNNs) require training and architectures for each new dataset, and compared to other fields, the potential of TabPFN to generalize to graph node classification remains an untapped area in graph learning.

\paragraph{Motivation 1: Limitations of LLM-dependent graph models.}
Recent graph foundation models fundamentally rely on LLMs to process node features~\citep{li2024glbench,liu2024ofa}. This dependency restricts them to text-attributed graphs where each node must have meaningful textual descriptions~\citep{he2024harnessing,li2024glbench}. Or they need textual instruction descriptions for prompt engineering~\citep{chai2023graphllm,ye2024language,tang2024graphgpt,chen2024llaga}.
Due to relying on LLMs, they require effort to create such textual descriptions, and some graph networks contain nodes with numerical features. Moreover, LLM-based approaches can introduce potential biases from pretrained language models. The field needs graph learning methods that handle arbitrary feature types without relying on language models.

\paragraph{Motivation 2: Success of Tabularization in Time-Series}
TabPFN offers an alternative paradigm. By training on millions of synthetic tabular datasets, it learns general classification patterns that transfer to real data without fine-tuning. TabPFN-TS~\citep{hoo2024tabular,hoo2025tables} recently demonstrated that this capability extends to time series by encoding temporal patterns into tabular format (see Table~\ref{tab:feature_analogy}), achieving competitive forecasting performance. This success demonstrates that structured domains can be ``tabularized'' via appropriate feature engineering for TabPFN. As shown in Table~\ref{tab:feature_analogy}, we propose an analogous transformation for graphs: extracting local structural patterns, global network properties, and positional encodings.

These motivations lead us to question: \emph{``Can we tabularize graph information into tabular features such that TabPFN can achieve competitive performance with GNNs without graph-specific training or LLM dependency?''}

\begin{figure}[t]
    \centering
    \includegraphics[width=0.85\linewidth]{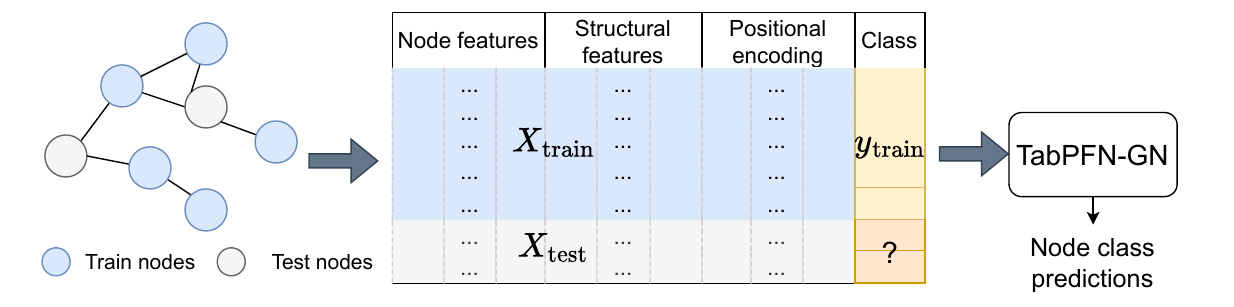}
    \caption{TabPFN-GN overview. Graph nodes are transformed into tabular features with node attributes, structural properties, and positional encodings, enabling direct inference via TabPFN.}
    \label{fig:overview}
    \vspace{-0.5em}
\end{figure}

We propose TabPFN for graph node classification (\textbf{TabPFN-GN}), which systematically transforms graph data into tabular representations for direct node classification as shown in Fig.~\ref{fig:overview}.
By encoding node attributes, structural properties, positional encodings, and optionally smoothed neighborhood features as tabular features, we enable TabPFN for node classification. Our experiments demonstrate that TabPFN-GN achieves competitive performance with GNNs on homophilous graphs and consistently outperforms them on heterophilous datasets, where the flexibility to exclude neighborhood aggregation proves advantageous. This success validates that principled tabularization can effectively capture graph structure.

\begin{table}[t!]
    \small
    \centering
    \caption{Analogous feature tabularization strategies.}
    \begin{tabular}{cll}
    \toprule
    Feature Type & TabPFN-TS~\citep{hoo2024tabular,hoo2025tables} & \textbf{TabPFN-GN} (Ours)\\
    \midrule
    Local Patterns & Calendar features & Degree, clustering, triangles \\
    Global Patterns & Seasonal features & Centrality (betweenness, PageRank)\\
    Position & Temporal index, sine/cosine encoding & LapPE, RWSE \\
    Smoothing & Moving average & Linear graph convolution \\
    \bottomrule
    \end{tabular}
    \label{tab:feature_analogy}
\end{table}

\section{Preliminaries \& Related Work}
\paragraph{Prior-data Fitted Network for Tabular Data.}
TabPFNv1~\citep{hollmann2023tabpfn} presents a new paradigm via the prior-data fitted network (PFN). It trains a transformer on millions of synthetic tabular data for in-context learning. This pre-training enables direct inference on small, real-world tabular data by leveraging the learned prior knowledge. 
TabPFNv2~\citep{hollmann2025tabpfnv2} extends this approach to handle larger datasets. For convenience, we will refer to both TabPFNv2 and TabPFNv1 as TabPFN.
To our knowledge, the only attempt to apply TabPFN to other domains is for time series forecasting. TabPFN-TS~\citep{hoo2024tabular,hoo2025tables} analyzes time series via feature engineering and encodes temporal patterns as tabular features. This success motivates our exploration of graph-to-tabular transformation.

\paragraph{Graph Neural Networks for Node Classification.}
While GNNs~\citep{kipf2017GCN,hamilton2017graphSAGE,velickovic2018GAT,choi2023gread} remain competitive on various graph tasks, they require dataset-specific training and architecture selection. Additionally,  neighborhood aggregation of GNNs shows stable performance on homophily graph benchmark datasets but struggles with heterophilous graphs~\citep{pei2020geom}. As GNNs may not dominate all graph networks, leveraging pretrained models such as TabPFN can bypass the need for architecture search. At the same time, we aim to verify their potential for node classification.

\paragraph{Graph Foundation Models}
Recent graph foundation models leverage LLMs. GraphGPT~\citep{tang2024graphgpt}, GraphLLM~\citep{chai2023graphllm}, and LLAGA~\citep{chen2024llaga} convert graphs to text descriptions, while frameworks that use text-attributed graph datasets~\citep{li2024glbench}, such as OFA~\citep{liu2024ofa}, use LLMs to encode node features. These approaches leverage LLMs' strengths and limitations, including their dependency on textual attributes. In contrast, our approach requires no LLMs and works with arbitrary node features.

\section{Proposed Method}

\subsection{Graph Tabularization for TabPFN}\label{sec:feature}
We transform graph structure and attributes into tabular features: original node features, structural features capturing connectivity patterns, positional encodings providing topological context, and optionally smoothed features from neighborhood aggregation (see Table~\ref{tab:feature_analogy}).

\paragraph{Node Attributes.}
We preserve original node features when dimensionally feasible. For high-dimensional features that do not satisfy the constraints of TabPFNs, we apply a truncated singular value decomposition (SVD) to preserve discriminative information.

\paragraph{Structural Features.}
We capture graph topology at local and global scales. Local structural features include degree, clustering coefficient, and triangle (i.e., 3-clique) count to quantify neighborhood patterns. Global structural features consist of centrality measures (e.g., betweenness~\citep{brandes2001centralitybet}, PageRank~\citep{page1999pagerank}) that encode network-wide importance.

\paragraph{Positional Encodings.}
In this study, we use either the LapPE~\citep{rampavsek2022gps} or the RWSE~\citep{rampavsek2022gps} as features. LapPE uses the first $k$ eigenvectors of the graph Laplacian to provide spectral coordinates. RWSE computes landing probabilities to encode multi-scale proximity relationships. These encodings distinguish structurally different nodes with similar attributes. More details are provided in Appendix~\ref{app:pos}.

\paragraph{Final Set of Features.}
Our final feature representation for each node $v$ combines complementary views extracted from the graph structure $\mathcal{G} = (\mathcal{V}, \mathcal{E})$ with normalized adjacency matrix $\bar{A}$:
\begin{equation}
   \mathbf{x}_v = [\phi_{\text{attr}}(v) \oplus \phi_{\text{struct}}(v,\bar{A}) \oplus \phi_{\text{pos}}(v,\bar{A}) \oplus \phi_{\text{smooth}}(v,\bar{A})],
\end{equation}
where $\phi_{\text{attr}}(v)$ represents the raw node features, $\phi_{\text{struct}}(v, \bar{A})$ captures both local patterns (degree, clustering coefficient, triangle count) and global importance (betweenness, closeness, PageRank) computed from the adjacency matrix, $\phi_{\text{pos}}(v, \bar{A})$ combines Laplacian PE and Random Walk SE derived from the graph Laplacian, and $\phi_{\text{smooth}}(v, \bar{A})$ optionally aggregates features from neighboring nodes through $L$-step linear graph convolutions~\citep{Wu2019SGC} without any weight matrices. This tabularization preserves essential graph information while enabling direct inference through TabPFN.

\subsection{Node Classification with TabPFN}
We directly input the features described in Sec.~\ref{sec:feature} into TabPFN for classification. 
Given training nodes with their tabularized features $X_{\mathrm{train}}=\{\mathbf{x}_i\}_{i\in \mathcal{V}_{\mathrm{train}}}$ and labels $y_{\mathrm{train}}=\{y_i\}_{i\in \mathcal{V}_{\mathrm{train}}}$, 
TabPFN performs in-context inference by learned patterns during pretraining.
For each test node $v \in \mathcal{V}_{\mathrm{test}}$ with features $\mathbf{x}_v$, TabPFN outputs an approximate posterior predictive distribution $p(y | X_{\mathrm{train}}, y_{\mathrm{train}}, \mathbf{x}_v)$, providing node-specific calibrated class probabilities without training. 
We follow the standard procedure of TabPFNv2 to apply z-normalization to all features.
All other configurations are left at their default values.

\section{Experiments}
\paragraph{Datasets.} 
We use both homophily and heterophily graph benchmark datasets for node classification. For homophily graph datasets~\citep{kipf2017GCN,rozemberczki2021multi,shirzad2023exphormer}, we use Cora, Citeseer, Pubmed, WikiCS, Amazon-Computer, and Amazon-Photo. For heterophily graph datasets~\citep{pei2020geom,platonov2023critical}, we compare GNNs on Chameleon, Squirrel, Cornell, Texas, Actor, and Wisconsin.
\paragraph{Evauluation Protocol.}
For Cora, Citeseer, and Pubmed, we follow the semi-supervised setting of \citet{kipf2017GCN} for data splits. We adhere to the widely accepted practice of training/validation/test splits of 60\%/20\%/20\% and the accuracy metric~\citep{shirzad2023exphormer,deng2024polynormer}. Furthermore, we utilize the WikiCS dataset and the splits provided in~\citet{rozemberczki2021multi}.
For Chameleon and Squirrel, we use the splits from \citet{platonov2023critical}, and for the other heterophilous datasets, we use the splits from \citet{pei2020geom}. More detailed settings are provided in Appendix~\ref{app:setting}.

\paragraph{Baseline.} 
We compare against standard GNNs (GCN~\citep{kipf2017GCN}, GraphSAGE~\citep{hamilton2017graphSAGE}, GAT~\citep{velickovic2018GAT}), GraphGPS~\cite{rampavsek2022recipe}, which combines Graph Transformers with local GNNs, and specialized heterophilous models (H2GCN~\citep{zhu2020beyond}, GPRGNN~\citep{chien2021GPRGNN}). For GraphAny~\citep{zhao2025graphany}, we use their arxiv-pretrained checkpoint for heterophilous datasets. For fair comparison, we re-evaluate GraphAny on Cora, Citeseer, and PubMed using our experimental setup.
\paragraph{Empirical Comparison.}
As shown in Table~\ref{tab:result_homo}, TabPFN-GN achieves competitive performance on homogeneous benchmarks, ranking first on Pubmed, WikiCS, and Computer.
As shown in Table~\ref{tab:result_hetero}, for heterogeneous graphs, TabPFN-GN achieves the best performance in all cases except Cornell. Furthermore, it consistently outperforms specialized models designed for such graphs, such as H2GCN and GPRGNN, except for Cornell. In particular, TabPFN-GN outperforms vanilla TabPFN on all datasets and consistently outperforms GraphAny.
\begin{table}[t!]
    \footnotesize
    \centering
    \caption{Test accuracy on homophilous graph benchmarks. \textbf{Best} and \underline{second-best} are highlighted.}
    \label{tab:result_homo}
    \begin{tabular}{l ccc ccc }\toprule
        Dataset & Cora & Citeseer & Pubmed & WikiCS & Computer & Photo\\\midrule
        GCN  & 81.60\std{0.40} & 71.60\std{0.40} & 78.80\std{0.60} & 77.47\std{0.85} & 89.65\std{0.52} & 92.70\std{0.20} \\ 
        GraphSAGE  & 82.68\std{0.47} & 71.93\std{0.85} & 79.41\std{0.53} & 74.77\std{0.95} & 91.20\std{0.29} & \underline{94.59\std{0.14}} \\ 
        GAT  & \textbf{83.00\std{0.70}} & 72.10\std{1.10} & 79.00\std{0.40} & 76.91\std{0.82} & 90.78\std{0.13} & 93.87\std{0.11} \\ 
        \cmidrule(lr){1-7}
        GraphGPS & \underline{82.84\std{1.03}} & \textbf{72.73\std{1.23}} & \underline{79.94\std{0.26}} & \underline{78.66\std{0.49}} & \underline{91.19\std{0.54}} & \textbf{95.06\std{0.13}} \\ 
        \cmidrule(lr){1-7}
        TabPFN & 57.30\std{0.00} & 51.50\std{0.00}  & 65.30\std{0.00} & 72.08\std{0.59} & 76.70\std{0.00} & 93.27\std{0.00} \\ 
        \cmidrule(lr){1-7}
        GraphAny & 79.38\std{0.16} & 68.10\std{0.04}  & 76.30\std{0.09}  & 74.95\std{0.61} & 83.04\std{1.24} & 90.60\std{0.82} \\ 
        \cmidrule(lr){1-7}
        \textbf{TabPFN-GN} & 81.98\std{0.45} & \underline{72.14\std{0.58}}  & \textbf{82.74\std{0.10}} & \textbf{79.40\std{0.77}} & \textbf{92.71\std{0.03}} & 93.55\std{0.05} \\ 
    \bottomrule
    \end{tabular}
\end{table}

\begin{table}[t!]
    \footnotesize
    \centering
    \caption{Test accuracy on heterophilous graph benchmarks.}
    \label{tab:result_hetero}
    \begin{tabular}{l ccc cc c}\toprule
        Dataset & Chameleon & Squirrel & Cornell & Texas & Actor & Wisconsin\\\midrule
        GCN  & 41.31\std{3.05} &  38.67\std{1.84} & 43.78\std{3.15} & 59.73\std{9.70} & 25.87\std{1.21} & 47.65\std{6.20}\\ 
        GraphSAGE  & 37.77\std{4.14} &  36.09\std{1.99} & 70.73\std{6.59} & 60.20\std{7.21} & 31.24\std{1.71} & 41.15\std{5.65}\\ 
        GAT  & 39.21\std{3.08} &  35.62\std{2.06} & 54.60\std{7.90} & 60.54\std{6.22} & 27.82\std{0.28} & 44.31\std{8.16}\\ 
        \cmidrule(lr){1-7}
        H2GCN  & 26.75\std{3.64} &  35.10\std{1.15} & 71.62\std{5.57} & \underline{79.73\std{3.25}} & \underline{36.18\std{0.45}} & 77.57\std{4.11}\\ 
        GPRGNN  & 39.93\std{3.30} &  \underline{38.95\std{1.99}} & \textbf{80.27\std{8.11}} & 78.38\std{4.36} & 35.30\std{0.80} & \underline{82.66\std{5.62}}\\ 
        \cmidrule(lr){1-7}
        TabPFN & \underline{45.16\std{4.32}} & 37.51\std{1.27}  & 72.70\std{6.33} & 79.19\std{3.83} & 36.37\std{1.31} & 82.55\std{4.15}\\ 
        \cmidrule(lr){1-7}
        GraphAny & 39.98\std{3.12} & 38.74\std{2.01} & 65.94\std{1.48} & 72.97\std{2.71} & 28.60\std{0.21} & 71.77\std{5.98}\\ 
        \midrule
        \textbf{TabPFN-GN} & \textbf{49.11\std{4.34}} & \textbf{46.66\std{1.43}} & \underline{74.05\std{6.96}} & \textbf{80.81\std{4.75}} & \textbf{37.22\std{1.08}} & \textbf{85.10\std{4.66}}\\ 
    \bottomrule
    \end{tabular}
\end{table}

\section{Discussion and Conclusion}
\paragraph{Limitations.}
TabPFN's constraint on class numbers prevents application to datasets like ogbn-arxiv (40 classes)~\citep{hu2021ogb}. While TabPFN-GN excels on heterophilous graphs, the synthetic prior lacks explicit graph connectivity patterns, potentially limiting performance on strongly homophilous networks. Future work should explore pre-training with graph-aware synthetic datasets and comprehensive comparisons with LLM-based graph foundation models.
\paragraph{Conclusion.}
We introduced TabPFN-GN, demonstrating that graph node classification can be effectively reformulated as tabular learning via principled feature engineering --- combining positional encodings, structural features, node attributes, and optional neighborhood aggregation. 
This tabularization achieves competitive performance without graph-specific training, particularly on heterophilous graphs.

\bibliographystyle{unsrtnat}
\bibliography{reference}
\clearpage
\appendix
\section{Dataset Statistics}
We list the dataset statistics we used in Tables~\ref{tab:data_homo} and~\ref{tab:data_hete}.
\begin{table}[h]
    \centering
    \small
    \caption{ Homophily dataset statistics for node classification benchmarks.}
    \label{tab:data_homo}
    \begin{tabular}{lcccccc}
    \toprule
     & Cora & Citeseer & Pubmed & Compuiter & Photo & WikiCS \\
    \midrule
    \#Nodes   & 2,708 & 3,327 & 19,717 & 13,752 & 7,650 & 11,701\\
    \#Edges   & 5,278 & 4,676 & 44,327 & 245,861 & 119,081 & 216,123\\
    \#Features &1,433 & 3,703 & 500 & 767 & 745 & 300\\
    \#Classes &  6 & 7 & 3 & 10 & 8 & 10\\
    \bottomrule
    \end{tabular}
\end{table}
\begin{table}[h]
    \centering
    \small
    \caption{Heterophily dataset statistics for node classification benchmarks.}
    \label{tab:data_hete}
    \begin{tabular}{lcccccc}
    \toprule
     & Texas & Wisconsin & Cornell & Actor & Squirrel & Chameleon  \\
    \midrule
    \#Nodes & 183 & 251 & 183 & 7,600 & 2,223 & 890 \\
    \#Edges & 295 & 466 & 280 & 26,752 & 46,998 & 8,854 \\
    \#Features & 1,703 & 1,703 & 1,703 & 931 & 2,089 & 2,325 \\
    \#Classes & 5 & 5 & 5 & 5 & 5 & 5 \\
    \bottomrule
    \end{tabular}
\end{table}
\section{Positional Encodings}\label{app:pos}
\paragraph{Laplacian Positional Encoding (LapPE).} 
Given a graph $\mathcal{G} = (\mathcal{V}, \mathcal{E})$ with adjacency matrix $A$ and degree matrix $D$, the normalized graph Laplacian is defined as:
\begin{equation}
    L = I - D^{-1/2}AD^{-1/2} = U\Lambda U^T,
\end{equation}
where $U = [u_1, u_2, ..., u_n]$ contains orthonormal eigenvectors and $\Lambda$ is the diagonal matrix of eigenvalues $0 = \lambda_1 \leq \lambda_2 \leq ... \leq \lambda_n \leq 2$.
LapPE~\citep{rampavsek2022recipe} uses the first $k$ non-trivial eigenvectors as positional features for node $v$:
\begin{equation}
    \text{LapPE}(v) = [u_2(v), u_3(v), ..., u_{k+1}(v)] \in \mathbb{R}^k.
\end{equation}
These eigenvectors provide a spectral coordinate system where geometrically close nodes have similar encodings. 

\paragraph{Random Walk Structural Encoding (RWSE).}
RWSE~\citep{rampavsek2022recipe} encodes the probabilities of random walks starting from each node. Let $P = D^{-1}A$ be the transition matrix. The probability of a random walk from node $v$ returning to itself in exactly $i$ steps is:
\begin{equation}
    p_i(v) = [P^i]_{v,v} = \text{diag}(P^i)[v].
\end{equation}
RWSE computes these probabilities for walks of different lengths:
\begin{equation}
    \text{RWSE}(v) = [p_1(v), p_2(v), ..., p_k(v)] \in \mathbb{R}^k.
\end{equation}
This encoding captures multi-scale structural information: $p_1(v)$ reflects immediate neighborhood density (related to degree), while larger $i$ values capture broader topological patterns and community structures.

\section{Detailed Experimental Settings}\label{app:setting}
\paragraph{Hardware and Software Specifications.}
Our implementation is based on \textsc{PyG} and \textsc{TabPFN}.
We run the experiments on a single \textsc{NVIDIA} RTX A6000 GPU with \textsc{CUDA} 12.4, \textsc{NVIDIA} Driver 550.54.14, and an i9 CPU.
\paragraph{Hyperparameters Configurations.}
We conducted experiments with the following hyperparameter search space:
\begin{itemize}
    \item Truncated SVD dimensions: $\{$None, 16, 32, 64, 128, 256$\}$
    \item PE type: $\{$Laplacian PE, Random Walk SE$\}$
    \item Dimensions of PE: $\{4, 8, 16, 32, 64\}$
    \item Local structural features: Degree, clustering coefficient, triangle count
    \item Global structural features: Betweenness centrality, PageRank
    \item $L$-steps linear graph convolutions: $\{0, 1, 2\}$
\end{itemize}
\paragraph{Feature Selection Strategy.}
Our framework allows flexible feature combination --- we can use all feature types comprehensively or selectively choose subsets based on dataset characteristics. For datasets with sufficient original node features (e.g., Citeseer with 3,703 features), we do not apply truncated SVD for dimensionality reduction, preserving the full feature information. For heterophilous datasets where neighborhood aggregation assumptions are violated, we exclude smoothed features from linear graph convolutions (i.e., set $L=0$). 
\paragraph{TabPFN-GN Inference Protocol.}
For TabPFN-GN's inference interface, we strictly maintain the integrity of the train/validation/test split. Only the training nodes with their labels $(X_{\text{train}}, y_{\text{train}})$ are provided as context to TabPFN-GN. The validation and test nodes are treated as query nodes, with TabPFN-GN predicting their labels based only on the training context. We ensure no label leakage by never exposing validation or test labels during inference, maintaining fair comparison with supervised GNN baselines that follow the same data split protocol.

\section{Additional Related Work}
Recent advances in graph foundation models have explored various directions toward generalizable and zero-shot graph learning.
AnyGraph~\citep{xia2024anygraph} addresses the challenge of distribution shifts in graph data by employing a Mixture-of-Experts (MoE) architecture with dynamic routing, resulting in strong zero-shot performance and fast adaptation to new datasets.
GCOPE~\citep{zhao2024all} enables unified pretraining across multiple graph domains by linking datasets with learnable coordinator nodes and aligning features via SVD, which mitigates the negative transfer of isolated pretraining and yields strong few-shot node-classification transfer.
The TS-GNNs framework~\citep{finkelshtein2025equivariance} introduces a recipe for building graph foundation models based on a ‘triple-symmetry’ principle: equivariance to node and label permutations, and invariance to feature permutations, thereby achieving strong zero-shot generalization across diverse datasets.

Several studies have explored language–graph integration~\citep{li2024glbench}.
\citet{ye2023instructglm} proposed InstructGLM, a framework that represents graph structures through flexible, scalable natural language descriptions. Instruction-finetuning an LLM with these descriptive prompts demonstrates superior performance over traditional GNN baselines on node classification and link prediction.
GraphText~\citep{zhao2023graphtext}  translates graphs into natural language by constructing a graph-syntax tree over node attributes and relationships and then traversing it to produce a textual prompt that supports training-free reasoning via in-context learning, while also being adaptable to instruction-tuning.
LLaGA~\citep{chen2024llaga} integrates LLMs with graph data by reorganizing nodes into structure-aware sequences and mapping them into the token embedding space with a versatile projector. This approach allows a single general-purpose model to achieve strong performance across various graph tasks and datasets, even outperforming specialized GNNs in both supervised and zero-shot scenarios.
One for All (OFA) framework~\citep{liu2024ofa} trains a single GNN by unifying cross-domain graphs as text-attributed graphs and standardizing node, link, and graph tasks via a nodes of interest subgraphs and their prompt nodes. It also introduces a novel graph prompting paradigm that enables in-context learning, allowing the model to achieve few-shot and zero-shot capabilities without requiring fine-tuning.

In recently, there are two concurrent works\citet{eremeev2025turning,hayler2025graphs} that share conceptual closeness to our TabPFN-GN by reformulating graph learning as tabular inference.

\section{Additional Studies}

\subsection{Comparison with LLM-based Graph Methods on GLBench}
Following the experimental setting of recent LLM-based graph methods, we conduct the supervised node classification experiments on all the datasets in GLBench~\citep{li2024glbench}~\footnote{\url{https://github.com/NineAbyss/GLBench}}.

TabPFN-GN achieves competitive or superior performance compared to LLM-based graph foundation models without requiring text descriptions or language model dependencies. While LLM-based methods leverage pre-trained language knowledge, TabPFN-GN leverages pre-trained patterns from massive synthetic prior data.

\begin{table}[h!]
    \footnotesize
    \centering
    \caption{Accuracy under the supervised setting of GLBench~\citep{li2024glbench}. \textbf{Best} and \underline{second-best} are highlighted.}
    \label{tab:glbench}
    \begin{tabular}{l cccc }\toprule
        Dataset & Cora & Citeseer & Pubmed & WikiCS \\\midrule
        InstructGLM~\citep{ye2023instructglm}     & 69.10 & 51.87 & 71.26 & 45.73 \\
        GraphText~\citep{zhao2023graphtext}       & \underline{76.21} & 59.43 & \underline{75.11} & 67.35 \\
        LLaGA~\citep{chen2024llaga}& 74.42 & 55.73 & 68.82 & 73.88 \\
        OFA~\citep{liu2024ofa} & 75.24 & \textbf{73.04} & \textbf{75.61} & \underline{77.34} \\
        \cmidrule(lr){1-5}
        \textbf{TabPFN-GN (Ours)}& \textbf{76.45} & \underline{63.33} & 66.74 & \textbf{77.72} \\
    \bottomrule
    \end{tabular}
\end{table}

\subsection{Comparison with tuned GNNs}
The GNNs in the main experiments do not use residual connections or other specific design options, as in~\citet{luo2024classic}. 
We compare the results of Chameleon and Squirrel using tuned GNNs (e.g., GCN$^\ast$, GraphSAGE$^\ast$, GAT$^\ast$) to compare the results of TabPFN-GN with those of~\citet{luo2024classic}.

\begin{table}[h!]
    \footnotesize
    \centering
    \caption{Compare with \citet{luo2024classic}'s setting}
    \label{tab:luo}
    \begin{tabular}{l cc}\toprule
        Dataset & Chameleon & Squirrel \\\midrule
        GCN$^\ast$ & 46.29\std{3.40} & 45.01\std{1.63} \\ 
        GAT$^\ast$ & 44.13\std{4.17} & 41.73\std{2.07} \\ 
        GraphSAGE$^\ast$  & 44.81\std{7.04} & 40.78\std{1.47}\\ 
        \cmidrule(lr){1-3}
        \textbf{TabPFN-GN} & 49.11\std{4.34} & 46.66\std{1.43} \\ 
    \bottomrule
    \end{tabular}
\end{table}

\subsection{Applicability to Graph Classification}
TabPFN-GN extends naturally to graph-level tasks by applying pooling operations (e.g., sum) over node features to obtain graph-level representations. We evaluate on the \textsc{IMDB-BINARY} (1,000 graphs), \textsc{Mutag} (188 graphs), and \textsc{Enzymes} (600 graphs), \textsc{Proteins} (1,113 graphs) tasks from TUDatasets~\citep{Morris2020TUDataset}. 
We report the TU datasets’ accuracy mean and STD of a 10-fold cross-validation runs in Table~\ref{tab:graph}.
We follow the same settings as~\citet{keren2024sequential} and use their results for the reported results.
In Table~\ref{tab:graph}, TabPFN-GN achieves the best performance on all 4 benchmarks, with particularly strong improvements on \textsc{Enzymes}.

\begin{table}[h!]
    \footnotesize
    \centering
    \caption{Results on TUDataset~\citep{Morris2020TUDataset}}
    \label{tab:graph}
    \begin{tabular}{l cccc }\toprule
        Dataset & \textsc{Proteins} & \textsc{Enzymes} & \textsc{Mutag} & \textsc{IMDB-Binary}\\\midrule
        GCN~\citep{kipf2017GCN} & 75.39\std{4.53} & 51.00\std{10.63} & 84.23\std{9.86} & 68.8\std{3.49} \\
        GAT~\citep{velickovic2018GAT} & 73.32\std{3.08} & 50.67\std{4.92} & 75.51\std{11.72} & 51.0\std{6.07} \\ 
        GIN~\citep{xu2018powerful} & 73.30\std{5.11} & 49.50\std{4.58} & 86.45\std{8.17} & 71.3\std{3.97} \\ 
        PNA~\citep{corso2020principal} & 74.86\std{4.57} & 52.50\std{4.60} & 84.19\std{9.44} & 71.9\std{4.46} \\
        \cmidrule(lr){1-5}
        TabPFN-GN & 76.80\std{3.78} & 61.43\std{5.76} & 88.36\std{6.07} & 73.3\std{3.83} \\
    \bottomrule
    \end{tabular}
\end{table}

\end{document}